# Robust Body Composition Analysis by Generating 3D CT Volumes from Limited 2D Slices

Lianrui Zuo[a], Xin Yu[b], Dingjie Su[b], Kaiwen Xu[c], Aravind R. Krishnan[a], Yihao Liu[a], Shunxing Bao[a], Fabien Maldonado[d,e], Luigi Ferrucci[f], and Bennett A. Landman[a,b]

[a]Department of Electrical and Computer Engineering, Vanderbilt University, Nashville, United States
[b]Department of Computer Science, Vanderbilt University, Nashville, United States
[c]Insitro, South San Francisco, United States
[d]Department of Medicine, Vanderbilt University Medical Center, Nashville, United States
[e]Department of Thoracic Surgery, Vanderbilt University Medical Center, Nashville, United States
[f]National Institute on Aging, National Institutes of Health, Baltimore, United States

**ABSTRACT**

Body composition analysis provides valuable insights into aging, disease progression, and overall health conditions. Due to concerns of radiation exposure, two-dimensional (2D) single-slice computed tomography (CT) imaging has been used repeatedly for body composition analysis. However, this approach introduces significant spatial variability that can impact the accuracy and robustness of the analysis. To mitigate this issue and facilitate body composition analysis, this paper presents a novel method to generate 3D CT volumes from limited number of 2D slices using a latent diffusion model (LDM). Our approach first maps 2D slices into a latent representation space using a variational autoencoder. An LDM is then trained to capture the 3D context of a stack of these latent representations. To accurately interpolate intermediate slices and construct a full 3D volume, we utilize body part regression to determine the spatial location and distance between the acquired slices. Experiments on both in-house and public 3D abdominal CT datasets demonstrate that the proposed method significantly enhances body composition analysis compared to traditional 2D-based analysis, with a reduced error rate from 23.3% to 15.2%.

## 1. INTRODUCTION

Body composition (BC) measures the proportion of muscle, bone, and fat of the human body, offering valuable insights into aging, disease progression, and overall human health conditions.[1,2] Studies have found strong connections between BC and cancer,[3,4] heart disease,[5] diabetes,[6] and sarcopenia.[7] BC can be estimated through various methods. Computed tomography (CT)[8] and magnetic resonance imaging[9] are widely used due to their high resolution and ability to differentiate between various tissue types. Dual-energy X-ray absorptiometry[10] is another common technique, especially for bone density and body composition assessment, although it provides less spatial details compared to CT and MRI. Additionally, bioelectrical impedance analysis and anthropometric measurements offer non-invasive and quick alternatives but with lower accuracy.[11]

Among all the methods for BC analysis, one widely used technique is CT body composition (CTBC).[1] Currently, the standard practice for CTBC analysis involves the use of two-dimensional (2D) single-slice CT scans. For example, the Baltimore Longitudinal Study of Aging (BLSA)[12] acquires two 2D slices of the human abdomen for CTBC analysis. Acquiring 2D slices is usually preferred over 3D CT to reduce unnecessary radiation exposure.[13] However, difficulty in consistently positioning cross-sectional locations leads to challenges in acquiring the 2D slices at the same desired axial position (vertebral level). As shown in Fig. 1, the ratio between subcutaneous fat and the muscle can vary significantly from slice to slice for 2D area-based calculations compared to a 3D volume-based calculation. Since the specific locations scanned are highly associated with BC measurements, increased positional variance in 2D abdominal slices complicates

---

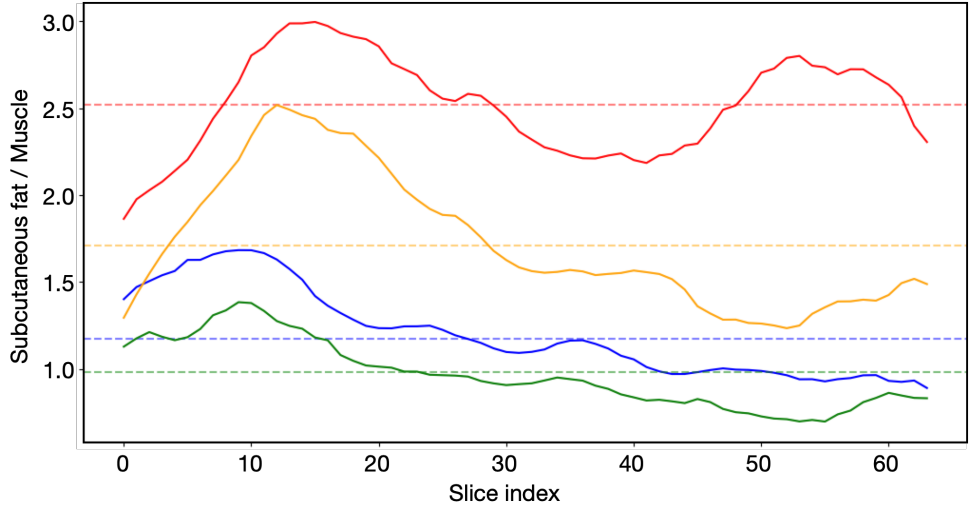

Figure 1. The ratio between abdominal subcutaneous fat and the skeletal muscle from four people. This ratio varies significantly from slice to slice, if it is calculated based on 2D area. Dotted lines show estimation based on 3D volumes.

BC analyses. This spatial variance issue is further compounded in longitudinal analysis, where tracking the BC change over time from single-slice CT images can be challenging.[13]

To address this, researchers proposed a 2D-2D image synthesis model, which analyzes the position of a 2D slice with respect to the human body and the uses a generative model to translate the slice to a target position while maintaining the underlying anatomical information.[13] However, as shown in Fig. 1, area-based BC estimation is sensitive to the localization of the 2D slice, leading us to argue that a 3D volume-based estimation model for BC analysis is preferred.

Considering the robustness of 3D volume-based BC analysis, we take a different route in this paper by investigating the possibility of "interpolating" and "extrapolating" the two acquired slices to construct a 3D volume for BC analysis. To draw an analogy, consider a tailor who only requires only a few measurements to create a well-fitting garment. Similarly, we hypothesize that a few acquired 2D CT slices, along with their relative locations in the human body, should provide rich information about BC. Instead of focusing on the exact anatomical details of organs, we infer a 3D volume based on the 2D slices and their estimated locations. Our approach includes two parts, a body part regression (BPR)[14] locating module and an image generating module based on latent diffusion models (LDMs). The BPR module takes 2D CT images as input and estimates the relative location of the image with respect to the human body. The image generation module then uses the acquired 2D CT images in conjunction with the BPR features to generate synthetic image slices to construct a 3D volume. We evaluated our method both qualitatively and quantitatively on held-out abdominal CT datasets. Results show that the volume-based BC analysis achieves a reduced error rate from 23.3% to 15.2%. In summary, the contributions of the paper are as follows: First, we propose an image generative model that synthesizes 3D volumes from only a few conditioned 2D slices. Second, we introduce a robust BC estimation tool for improved CTBC analysis given limited 2D data.

## 2. METHODS

### 2.1 Framework

The proposed method synthesizes 3D CT volumes from 2D slices for CTBC analysis. Our approach leverages two key pieces of information to guide the model towards better synthesis results. First the model is informed by the location of the 2D slices, especially considering variability in acquisition locations. Second, the model infers the anatomical distribution of the human abdomen to synthesize a likely volume based on the conditioning 2D slices. Accordingly, our proposed approach for BC analysis consists of two key components: the BPR module[14] and the LDM.[15] The high-level workflow of the method is illustrated in Fig. 2. The BPR module estimates the location of the acquired slices, while the LDM takes the BPR features along with the encoded latent representations of the acquired slices as input to generate a 3D image consisting of $N_{\text{total}}$ slices.

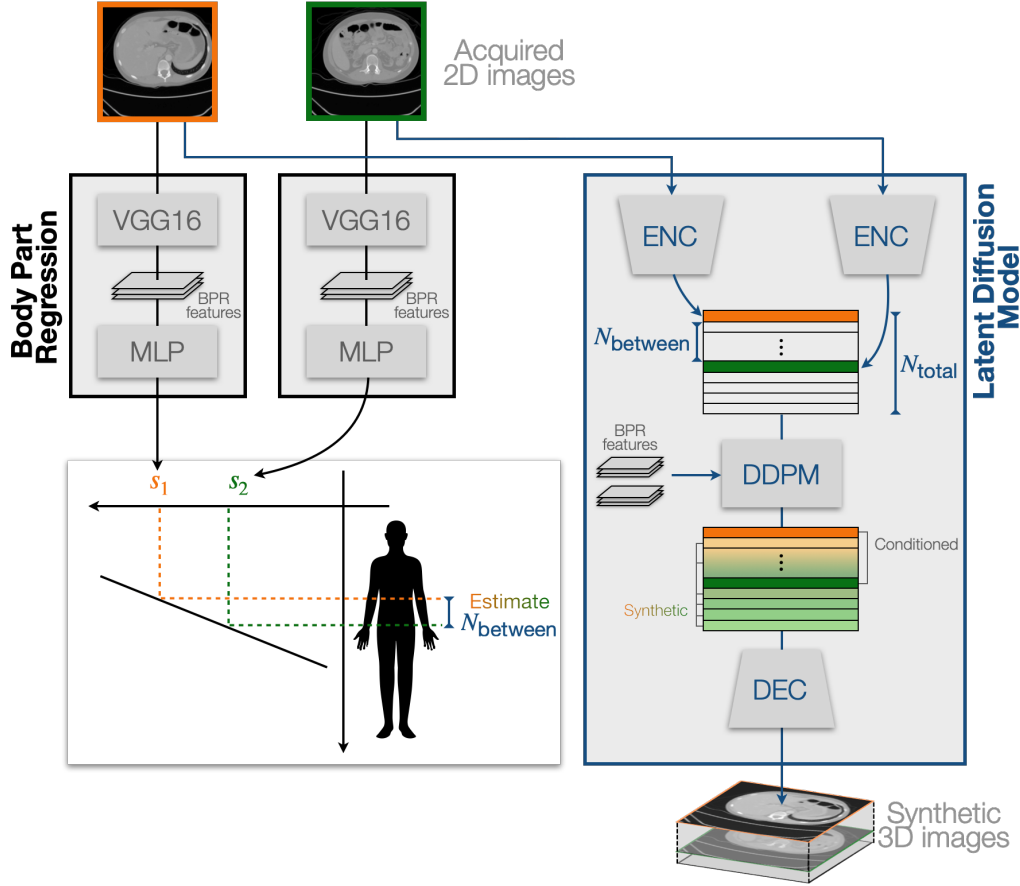

Figure 2. The diagram illustrates two key modules: Body part regression (BPR) and latent diffusion model (LDM). The BPR module estimates the location of the acquired 2D images by comparing the BPR score $s$ with a reference whole-body CT atlas. The LDM uses the BPR features long with the encoded representations of the acquired slices to generate a 3D image consisting of $N_{total}$ slices.

For each 2D CT slices, the BPR module predicts a score $s$ as its output. The absolute value of the score does not indicate the physical location of the slice; however, CT slices with similar scores are likely to be close to each other, even if they may come from different individuals. To estimate the physical location of an acquired slice, we first run BPR on a whole-body CT and obtain the score $s$ for each slice as a reference. During application, each acquired slice will be processed by BPR to find the best match between the testing image and the reference image in terms of their score.

During image synthesis, when two slices are provided as conditions, the BPR module first estimates the scores $s$ and selects the upper (superior) slice as the starting slice. It then estimates the location of the second slice relative to the starting slice. This relative location is converted into the number of slices $N_{between}$ that lie between the two acquired slices. When $N_{between}$ + 2 (including the starting and ending slices) is smaller than total number of slices in the synthetic image $N_{total}$, the LDM will generate additional slices below the second conditioning slice to have a total of $N_{total}$ slices in the synthetic 3D volume. This generation process involves both interpolation between the two acquired slices and extrapolation beyond the second slice to achieve a total of $N_{total}$ slices. When more than two slices are provided, the BPR module first selects the starting slice then uses all the remaining slices as conditions. These conditions help in accurately placing the intermediate slices and ensuring the continuity of anatomical structures. When only one slice is available, it is treated as the top slice of the $N_{total}$ estimated slices.

**2.2 Flexible Image Generation with LDM: Training and Inference**

Before training the proposed method, each 3D image in our abdominal CT dataset undergoes preprocessing. The images have dimensions of 512 × 512 × $S$, where $S$ is the number of axial slices, typically varying from image to image. Each CT image has a slice thickness of 3mm without a slice gap. To reduce computational load while preserving the anatomical details, we downsampled the images in the axial plane to $256 \times 256 \times S$ using Gaussian blurring as an anti-aliasing filer. The intensities of all images were then clipped to the range $[-1024, 3072]$ Hounsfield unites and normalized to $[-1,1]$. Finally, we randomly selected segments of consecutive $N_{total}$ slices per 3D image, resulting in a final dimension of $256 \times 256 \times N_{total}$ per data sample.

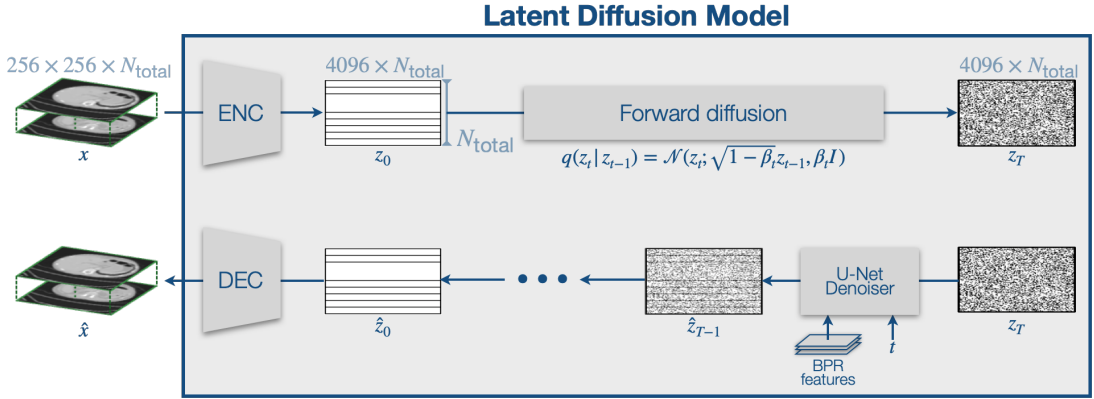

Figure 3. During training, the preprocessed images are first encoded into a latent space $z$. LDM is then trained on a stack of $z$'s from $N_{total}$ slices. During reverse diffusion process, BPR features from the top slices are also fed into the denoiser network to provide location information of the image slices.

We adopt the LDM proposed in Rombach et al.,[15] which includes a variational autoencoder (VAE)[16] and a denoising diffusion probabilistic model (DDPM),[17,18] as shown in Fig. 3. DDPMs have shown superior performance in realism and training stability than other generative models, such as generative adversarial networks (GANs).[15,17,24] To train the LDM, we first train a VAE that maps each 2D slice to a latent space $z$. This process maps each slice into a 4096-dimensional latent space, efficiently compressing the high-dimensional CT images while preserving essential features. Once the VAE is trained, the DDPM is trained in the VAE's latent space $z$. To teach the DDPM to capture the 3D contextual information, we stack the representations from $N_{total}$ slices. As discussed in Sec. 2.1, location information of each conditioning slices is crucial for the LDM to generate accurate latent representations of missing slices. Therefore, we feed both $z$ from the VAE and the feature maps from BPR into the DDPM during training. After training, the LDM can perform zero-shot inference without a retraining.

The core idea for zero-shot inference is inspired by the RePaint method.[19] Specifically, we modify the estimated latent representation $\hat{z}_t$ at each time step $t$ by incorporating the real $z_t$ values (with corresponding levels of noise) from the acquired slices:

$$\hat{z}_t \leftarrow z_t \circ M + \hat{z}_t \circ (1 - M), \tag{1}$$

where $M$ has the same dimension as $z$ and takes the value of 1 for positions corresponding to acquired slices and 0 for positions corresponding to unavailable slices. The operator "$\circ$" indicates element-wise multiplication. For slices that need to be predicted, the reverse diffusion process estimates the corresponding missing $z$ values, which are then sent to the VAE decoder to generate new slices. The $z$ values from the acquired data guide the diffusion process to fill in the missing pieces based on the learned prior anatomical knowledge.

### 2.3 Implementation Details

In our implementation, the total number of slices in the synthetic image is $N_{total} = 64$. Since our training data have slice thickness of 3mm, $N_s = 64$ covers approximately 20cm of the human body, providing 3D context for the LDM to capture. For the VAE, we utilized the implementation provided in MONAI 1.2.[20] The denoiser network $\epsilon_\theta(z_t, f_{BPR}; t)$ in our LDM is a U-Net[21] with four downsampling levels, where $f_{BPR}$ is the BPR feature. $f_{BPR}$ is zero for slices that do not have an associated BRP feature. We employed the cosine noise scheduling as recommended in the literature.[18] The number of diffusion steps is set to 1000. For training the VAE and LDM, we used data from $N = 323$ subjects from an abdomen dataset. All models were trained on a NVIDIA A6000 GPU with 48 GB of RAM.

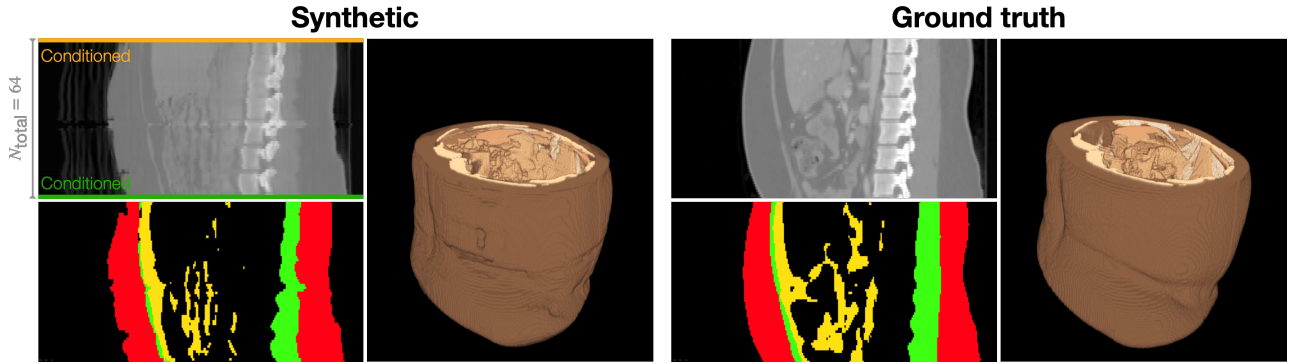

Figure 4. Left: synthetic 3D volume conditioned on a top and bottom slice with $N_{total} = 64$. Right: Acquired 3D abdominal CT with TotalSegmentator[22] results and a 3D rendering as ground truth. This image is randomly selected from the testing dataset.

Table 1. Error rate between the subcutaneous fat and the muscle, as well as the error rate between the viscera fat and the muscle are reported in the Table. Numbers are reported in "Mean ± Standard deviations".

|  | **Proposed (volume-based)** | **Top slice (area-based)** | **Bottom slice (area-based)** |
| --- | --- | --- | --- |
| $Err(R_{S.Fat})(\%)$ | 15.2 ± 7.3 | 23.3 ± 15.1 | 26.5 ± 16.2 |
| $Err(R_{V.Fat})(\%)$ | 26.3 ± 13.4 | 24.2 ± 12.9 | 25.4 ± 16.7 |

## 3. EXPERIMENTS AND RESULTS

To qualitatively and quantitatively evaluate the proposed method, we prepared a held-out dataset consisting of 20 3D abdominal CT images. Each image was preprocessed following the same preprocessing steps described in Sec. 2.2.

For each 3D image, we selected the center 64 axial slices as the ground truth 3D image and used the top and bottom slices as inputs to the proposed model to estimate BC. For qualitative evaluation, we visualize the generated CT images and their corresponding segmentation labels. We also provide 3D renderings of the segmentation labels to highlight the anatomical continuity and consistency of the generated volumes. As shown in Fig. 4, the synthetic image was generated by conditioning on a top slice and a bottom slice with $N_{total} = 64$. We then ran TotalSegmentator[22] to obtain the segmentation labels of subcutaneous fat (red), skeletal muscle (green), and visceral fat (yellow). It is encouraging to see that the subcutaneous fat and skeletal muscle generally align well with the ground truth labels from the acquired image. However, it is also worth noting that the proposed method does not generate accurate results of the organs inside the abdomen. We further elaborate on this limitation in Sec. 3.1.

### 3.1 Quantitative evaluation

We conduct a quantitative comparison between our proposed method (with two slices as conditioning) and traditional 2D slice-based methods for BC analysis. We report the error rate between the subcutaneous fat and the muscle $Err(R_{S.Fat})$, as well as the error rate between the visceral fat and the muscle $Err(R_{V.Fat})$. For example, $Err(R_{S.Fat})$ is defined as

$$Err(R_{S.Fat}) = \frac{|R_{S.Fat}(Syn) - R_{S.Fat}(GT)|}{R_{S.Fat}(GT)} \times 100\% \quad (2)$$

where $R_{S.Fat}$ is the volumetric ratio between the subcutaneous fat and the muscle. It is encouraging to see that the proposed method reduced the error rate of subcutaneous fat estimation from 23.3% for a 2D area-based method to 15.2%. However, the proposed method does not show improvements in estimation the ratio between visceral fat and the muscle. We hypothesize this is due to the limited performance of the proposed method in synthesizing abdominal organs.

### 3.2 Ablation studies

To investigate the impact of the number of conditioning slices on the performance of our method, we performed an ablation study by conditioning the proposed method on different numbers of input slices (1, 2, 3, and 4 slices) with $N_{total} = 64$. When there is only one slice as conditioning, it is used as the top slice. We analyzed the resulting BC estimation accuracy for each case to understand how additional slices influence the model's performance. Specifically, we focus on the prediction accuracy of the volumetric ratio between subcutaneous fat and the muscle, as well as the volumetric ratio between viscera fat and the muscle. Results are shown in the top two rows of Table 2. Both error rates of both ratios decrease as the number of conditioning slices increases. This makes sense since the more slices used as conditioning, the more information the model can use to predict the 3D volume.

Table 2. Ablation study to investigate the contribution of BPR features and the impact of different numbers of slices as conditioning. Numbers are reported in "Mean ± Standard deviations". Bold numbers indicate statistical significance ($\alpha = 0.05$) under Wilcoxon signed rank tests with and without using BPR features.

| Use BPR feature? | # of condition slices | 1 | 2 | 3 | 4 |
|---|---|---|---|---|---|
| Yes | $Err(R_{S.Fat})(\%)$ | **39.6±15.5** | **15.2±7.3** | 13.6±4.8 | 12.7±3.9 |
| Yes | $Err(R_{V.Fat})(\%)$ | 59.8±23.9 | 26.3±13.4 | 21.8±9.7 | 18.9±5.6 |
| No | $Err(R_{S.Fat})(\%)$ | 45.2±16.3 | 18.1±8.0 | 13.8±5.0 | 12.9±4.1 |
| No | $Err(R_{V.Fat})(\%)$ | 61.1±24.4 | 25.9±12.9 | 22.6±8.9 | 19.1±6.6 |

As discussed in Sec. 2.1, BPR features of the top slice is used to provide additional guidance to the DDPM about the location information. To evaluate the importance of this BPR feature in our proposed method, we compared the results with and without BPR guidance. This experiment involves running the proposed method under two settings: one with BPR guidance and one without (using only 2D slice information). We analyzed the impact of BPR guidance on the error rate of the generated 3D volumes and BC analysis. Results are shown in the bottom two rows of Table 2. In general, both error rates increase after removing the BPR features as guidance. We then conducted a Wilcoxon signed rank test between the error rates with and without using BPR features as guidance. Statistically significant results ($\alpha = 0.05$) are shown in bold numbers in Table 2. It is interesting that the BPR features are helpful in estimating the ratio between subcutaneous fat and the muscle when the number of conditioning slices is limited (i.e., one or two). When there are more slices available as conditioning, BPR features are no longer helpful.

## 4. DISCUSSION AND CONCLUSION

In this study, we investigated the generation of 3D CT volumes from limited 2D slices for BC analysis. Our proposed method integrates a BPR module and an LDM to synthesize anatomically consistent 3D volumes from 2D CT slices. The BPR module estimates the relative location of the slices within the human body, while the LDM leverages this spatial information to generate intermediate and extrapolated slices, forming a complete 3D volume. This approach aims to mitigate the spatial variance issues associated with single-slice methods and enhance the accuracy and robustness of CTBC analysis.

Despite the promising results, our study has several limitations. First, while the generated images provide more accurate BC analysis, the internal organs in these synthetic images may not be anatomically correct. This limitation highlights the need for future improvements, especially in estimating the viscera fat regions. Second, there are potential alternative approaches that warrant exploration. For example, directly generating 3D results based on segmentation labels rather than intensity images might offer a more direct and potentially more robust method for BC estimation. Third, the encouraging results of this study prompt us to consider expanding the current method to Dual-energy X-ray absorptiometry, which is also common for BC analysis. Specifically, we could investigate the potential of synthesizing a 3D volume from 2D projection scans, further broadening the applicability of our approach. Finally, identifying the application boundary of our approach---i.e., to which extend can a 3D volume be reliably generated from 2D slices---warrants further research. We hypothesize that this application boundary can be estimated using mutual information-based metrics. [23]

In conclusion, this study presents a novel approach for generating 3D CT volumes from limited 2D slices to enhance the accuracy and robustness of BC analysis. By leveraging the spatial information provided by the BPR module and the generative capabilities of the LDM, our approach provides a new perspective for BC analysis and overcomes the limitations of traditional 2D slice-based methods.

## ACKNOWLEDGMENTS

This research was funded by the National Cancer Institute (NCI) grants R01 CA253923 (Landman) and R01 CA275015 (Landman). This research was also supported by the NIH grants U54DK134302 and U54EY032442. The Vanderbilt Institute for Clinical and Translational Research (VICTR) is funded by the National Center for Advancing Translational Sciences (NCATS) Clinical Translational Science Award (CTSA) Program, Award Number 5UL1TR002243-03. The content is solely the responsibility of the authors and does not necessarily represent the official views of the NIH. This work was conducted in part using the resources of the Advanced Computing Center for Research and Education at Vanderbilt University, Nashville, TN.

**Declaration of the usage of generative AI:** In the creation of this work, generative AI technologies have been used with a focused and specific purpose: to assist in structuring sentences and performing grammatical checks. It is imperative to highlight that the conceptualization, ideation, and all prompts provided to the AI originate entirely from the author(s)'s creative and intellectual efforts. The AI's function in this process has been akin to that of a sophisticated, programmatically driven editorial tool, without any involvement in the generation of original ideas or content.